\documentclass{article}
\usepackage{graphicx} % Required for inserting images
\usepackage{parskip}
\usepackage{xparse}
\usepackage{xcolor}
\usepackage{amsmath}
\usepackage{amssymb}
\usepackage{amsthm}
\newtheorem{theorem}{Theorem}

\title{Jeffrey's update rule as a minimizer of Kullback-Leibler divergence}
\author{Carlos Pinzón \and Catuscia Palamidessi}
\date{February 2025}

\begin{document}

\maketitle
\begin{abstract}
In this paper, we show a more concise and high level proof than the original one, derived by researcher Bart Jacobs, for the following theorem: in the context of Bayesian update rules for learning or updating internal states that produce predictions, the relative entropy between the observations and the predictions is reduced when applying Jeffrey's update rule to update the internal state.
\end{abstract}

\NewDocumentCommand{\DKL}{s O{} O{}}{
    \IfBooleanTF{#1}
        {{D_{\text{KL}}}}
        {{\DKL*(#2\,||\,#3)}}
}
\NewDocumentCommand{\Cfwd}{}{
    \overrightarrow{C}}
\NewDocumentCommand{\Cbwd}{O{\ensuremath{\px}}}{
    \overleftarrow{C_{#1}}}
\NewDocumentCommand{\domX}{}{{\mathcal{X}}}
\NewDocumentCommand{\domY}{}{{\mathcal{Y}}}
\NewDocumentCommand{\xn}{}{{x_{1..n}}}
\NewDocumentCommand{\yn}{}{{y_{1..n}}}
\NewDocumentCommand{\oo}{}{{\infty}}
\NewDocumentCommand{\px}{}{{\ensuremath{\theta}}}
\NewDocumentCommand{\py}{}{{\ensuremath{\tau}}}
\NewDocumentCommand{\ppx}{}{{p_\px}}
\NewDocumentCommand{\pxt}{}{{\px_t}}
\NewDocumentCommand{\pxtt}{}{{\px_{t+1}}}
\NewDocumentCommand{\ppxt}{}{{p_{\pxt}}}
\NewDocumentCommand{\ppxtt}{}{{p_{\pxtt}}}

\section{Introduction}

In this paper, we show a more concise and high level proof of a theorem presented in~\cite{jacobs2021learning}, which says that in the context of Bayesian update rules for learning or updating beliefs about parameters, Jeffrey's update rule reduces the relative entropy $\DKL[\py][\Cfwd(\px)]$ when updating $\px$ from some prior to a posterior distribution.

The proof presented here consists of putting together in a consistent and uniform way several facts in the context of the Expectation-Maximization (EM) algorithm~\cite{elsalamouny2020generalized,dempster1977maximum,mclachlan2007algorithm,wu1983convergence}.

\section{Notation}

There is a channel $C(y|x)$ with input space $\domX=1..N$ and output space $\domY=1..M$, encoded as an $N\times M$ matrix $C_{x\,y}:=C(y|x)$ whose rows add up to one individually.
For $x\in \domX$, $C(x)$ denotes the distribution $y\mapsto C(y|x)$.
For distributions $\px$ over $\domX$, $\Cfwd(\px)$ denotes the push-forward distribution $\Cfwd(\px)(y) = \sum_x \px(x) C(y|x)$.
Analogously, for distributions $\py$ over $\domY$, $\Cbwd[](\py)$ denotes the likelihood function $x\mapsto C(y|x)$.
Provided a distribution $\px$, $\Cbwd[\px]$ denotes the inverse channel $\Cbwd[\px](x|y) := \px(x) C(y|x) / (\Cfwd(\px)(y))$, which can be used to map distributions $\py$ over $\domY$ to distributions $x\mapsto\Cbwd[\px](\py)(x)$ over $\domX$.

To keep the notation concise and semantically simple, since the channel $C$ is fixed, we remove the explicit dependence on $C$ in the following probabilities: $\ppx(y) := \Cfwd(\px)(y)$, $\ppx(x|y) := \Cbwd(y)(x)$ and $\ppx(x,y) := \ppx(y) \ppx(x|y) = \px(x) C(y|x)$.
However, for $\ppx(x)$ we write directly $\px(x)$, which is simpler, and for $\ppx(y|x)$ we write $p(y|x)$ or $C(y|x)$, which are simpler and make the irrelevance of $\px$ evident.

\section{Theorem statement}

In the problem under consideration, there is a known channel $C$, and a sequence of observations $\yn$, with $y_i\in\domY$, that correspond in order to an unknown sequence of inputs $\xn$, sampled i.i.d. from an unknown distribution $\px^\star$.
The given observations form an empirical distribution of frequencies $\py$ and the objective is to produce an estimate $\hat\px$ for $\px^\star$.
This objective is carried out in steps that repeatedly update some prior distribution $\pxt$ into some posterior distribution $\px_{t+1}$ until convergence $\hat\px:=\px_\oo$, starting with some fixed prior $\px_0$.

One such method for updating priors into posteriors is \textit{Jeffrey's update rule}, which is given by
\begin{equation}\label{eq:jeffreys-rule}
    \pxtt := \Cbwd[\pxt](\py)
\end{equation}
for distributions $\pxt$ with full-image, i.e. $\ppxt(y)>0$ for all $y\in \domY$.
This constraint for $\pxt$ is discussed and relaxed in a subsequent section, but for a quick and reassuring remark without entering into details, it suffices that $\px_0$ satisfies it and $C(y|x)>0$ for all $x,y$, to ensure that $\px_t$ satisfies it for all $t$.

The following theorem is proved by Bart Jacobs in~\cite{jacobs2021learning} using heavy machinery from linear algebra theory.
Our goal in this paper is to prove it with a simpler argument.

\begin{theorem}\label{eq:theorem}
Jeffrey's update rule reduces (or maintains at least) the relative entropy $\DKL[\py][\Cfwd(\px)]$.
\end{theorem}

\section{Proof}

\NewDocumentCommand{\DDKL}{O{}O{}}
{\ensuremath{\Delta}\DKL[#1][#2]}

Denote the log-likelihood of $\px$ (scaled by $1/n$) for the given observations $\yn$ (or $\tau$) as
\begin{equation}
    L(\px) := \frac{1}{n}\log \ppx(\yn) := \frac{1}{n}\log \prod_{i=1}^n \ppx(y_i) = \frac{1}{n}\sum_{i=1}^n \log \ppx(y_i) = \sum_y \py(y) \log \ppx(y),
    \label{eq:log-likelihood}
\end{equation}
with the convention $0\, \log 0 = 0$ for values of $y$ that are both unobserved and deemed impossible by $\px$.

For an update from $\px_t$ to $\px$, the log-likelihood difference (after minus before) $\Delta L(\px):= L(\px) - L(\pxt)$ is directly related with the KL-divergence difference via the following crucial observation:
\begin{align}
    \Delta L(\px) &:= L(\px) - L(\pxt)
    = \sum_{x,y} \tau(y) \log \frac{\ppx(y)}{\ppxt(y)}
    \notag\\
    &= \DKL[\py][\Cfwd(\px_{t})] - \DKL[\py][\Cfwd(\px)], \label{eq:DeltaL}
    \notag
    \text{ so}\\
    \Delta L(\px_{t+1}) \geq 0 &\text{ if and only if } \DKL[\py][\Cfwd(\px_{t+1})] \leq \DKL[\py][\Cfwd(\px_t)]
\end{align}

Therefore, it suffices to show that Jeffrey's update $\px_{t+1}$ satisfies $\Delta L(\px_{t+1})\geq 0$, a fact that is proved in the next section.
\hfill$\blacksquare$

\section{Jeffrey's rule in the EM algorithm}

In this section, we show why Jeffrey's update rule \eqref{eq:jeffreys-rule} increases the log-likelihood function \eqref{eq:log-likelihood}.
This is a well known fact from the theory of the Expectation-Maximization (EM) algorithm (see \cite{elsalamouny2020generalized,dempster1977maximum} for instance) and for the sake of completeness and clarity, we provide a complete proof using the notation of this paper.

Fix $\pxt$ and split the likelihood as $L(\px) = Q(\px|\pxt) + H(\px|\pxt)$, where
\begin{align}
    Q(\px|\pxt) &:= \frac{1}{n} \sum_{\xn} \ppxt(\xn|\yn) \log \ppx(\xn,\yn)
    \notag\\
    &= \sum_{x,y} \tau(y) \ppxt(x|y) \log \ppx(x,y),
    \\
    H(\px|\pxt) &:= -\frac{1}{n}\sum_{\xn} \ppxt(\xn|\yn) \log \ppx(\xn|\yn)
    \notag\\
    &= -\sum_{x,y} \tau(y) \ppxt(x|y) \log \ppx(x|y).
\end{align}

Notice indeed that $L(\px)=Q(\px|\pxt) + H(\px|\pxt)$ regardless of the value of $\pxt$ because $\log \ppx(x,y) - \log \ppx(x|y) = \log \ppx(y)$, an expression independent of $x$ just like $\tau(y)$, and since $\ppxt(x|y)$ is a distribution for $x$, then $\sum_x \ppxt(x|y)=1$, yielding
\begin{equation}
    Q(\px|\pxt) + H(\px|\pxt) = \sum_y \tau(y) \left(\sum_x \ppxt(x|y)\right) \log \ppx(y) = L(\px).
\end{equation}

Furthermore, we may decompose $\Delta L(\px)$ as $\Delta Q(\px) + \Delta H(\px)$ where
\begin{align}
    \Delta Q(\px) &:= Q(\px|\pxt) - Q(\pxt|\pxt)
    = \sum_{x,y} \tau(y) \ppxt(x|y) \log \frac{\ppx(x,y)}{\ppxt(x,y)}
    \notag\\
    &= \sum_{x,y} \tau(y) \ppxt(x|y) \log \frac{\px(x)}{\pxt(x)}
    = \sum_x \Cbwd[\pxt](\tau)(x) \log \frac{\px(x)}{\pxt(x)}
    \notag\\
    &= \DKL[\Cbwd[\pxt](\tau)][\px_{t}] - \DKL[\Cbwd[\pxt](\tau)][\px],\\
    \raisebox{2em}{\,}
    \Delta H(\px) &:= H(\px|\pxt) - H(\pxt|\pxt)
    = \sum_{y} \tau(y) \sum_x \ppxt(x|y) \log \frac{\ppxt(x|y)}{\ppx(x|y)}
    \notag\\
    &= \sum_{y} \tau(y) \DKL[\Cbwd[\pxt](1_y)][\Cbwd[\px](1_y)]
    \hspace{1.5em}(1_y\text{ is the indicator function)}
\end{align}

Since $\Delta H(\px)$ is an average of divergences, it is non-negative (a result known as Gibb's inequality), so it suffices to show that $\Delta Q(\px)\geq 0$ to conclude that $\Delta L(\px)\geq 0$.
This is not true in general for every $\px$, but it is for Jeffrey's posterior $\pxtt$ because the maximum of $Q(\px|\pxt)$ occurs precisely at $\px=\pxtt$ as shown next.

Applying the Lagrange multiplier method to the function $Q(\px|\pxt)$ with restriction $\sum_x \px(x)=1$, we obtain that $Q(\px|\pxt)$ is maximized when $\px$ satisfies
\begin{align}
    \forall x,\hspace{1em} 0
    &=\frac{\partial\; Q(\px|\pxt)-\lambda (\sum_{x'} \px(x')-1)}{\partial \px(x)}
    \notag\\
    &= \sum_{y} \tau(y) \ppxt(x|y) \frac{\partial\log \ppx(x,y)}{\partial \px(x)} - \lambda
    \notag\\
    &= \sum_{y} \tau(y) \ppxt(x|y) \frac{\ppx(y|x)}{\ppx(x,y)} - \lambda
    \notag\\
    &= \frac{1}{\px(x)} \sum_{y} \tau(y) \ppxt(x|y)  - \lambda
    \label{eq:lagrange}
\end{align}

From \eqref{eq:lagrange} and the constraint $\sum_x \px(x)=1$ it follows that $\lambda=\sum_{x,y} \tau(y) \ppxt(x|y) = \sum_{y} \tau(y) = 1$, and
\begin{equation}
    \px(x) = \frac{1}{\lambda} \sum_{y} \tau(y) \ppxt(x|y)
    = \Cbwd[\pxt](\tau)(x)
    = \pxtt(x). 
\end{equation}
\hfill$\blacksquare$

\section{Full-image constraint and sparsity}

Jeffrey's rule \eqref{eq:jeffreys-rule} is defined only for non-pathological prior distributions $\px$ for which $\ppx(y)>0$ for all $y$.
So far, we have ignored this detail in the proofs and assumed that if this condition holds for the initial prior $\px_0$, it will hold for all subsequent posteriors $\px_t$.
In this section we prove this assertion.

If $C(y|x)>0$ for all $x,y$, the result is immediate.
We will therefore consider the sparse case in which $C(y|x)=0$ for many combinations of $x$ and $y$.

Without loss of generality, it can be assumed that for every $x$, there is some $y$ with $\tau(y)>0$ for which $C(y|x)>0$.
Otherwise, since this $x$ contradicts the observed distribution $\tau$, it has zero chances of being an input and we can safely remove it from $\domX$ and analyze the channel without it.

\begin{align}
    \Cfwd(\pxtt)(y) &= \sum_x C(y|x) \pxtt(x)
    = \sum_x C(y|x) \Cbwd[\pxt](\tau)(x)
    \notag\\
    &= \sum_{x,y'} C(y|x) \tau(y') \ppxt(x|y')
    = \sum_{x,y'} C(y|x) \tau(y') \pxt(x) C(y'|x) / \ppxt(y')
    \notag\\
    &= \sum_{x,y'} \ppxt(x,y) \tau(y')  C(y'|x) / \ppxt(y')
\end{align}

Since $\ppxt(y)>0$, there are some $x$ with $\ppxt(x,y)>0$.
For each of these values $x$, there must exist some values $y'$ such that $\tau(y')C(y'|x)>0$ because of the argument in the previous paragraph.
Therefore, $\Cfwd(\pxtt)(y)>0$.
\hfill$\blacksquare$

Furthermore, we may consider a more relaxed constraint that allows distributions $\px$ for which $\ppx(y)>0$ whenever $\tau(y)>0$, so not necessarily for all $y$.
In other words, we consider distributions $\px$ for which $\yn$ is plausible (even if extremely unlikely) in the sense that $\ppx(y_i)>0$ for each $i=1..n$.
The argument for why it suffices to consider a well-behaved starting point to ensure well-behavior for all subsequent updates is exactly the same, but restricting the argument to values $y$ with $\ppx(y)>0$.

\section{Summary}

We provided an alternative proof to Theorem~\ref{eq:theorem} that is more simple and mathematically more elegant to the one in the state of the art~\cite{jacobs2021learning}.

First, we showed that the change in relative entropy $\DKL[\tau][\Cfwd(\px)]$ can be fully characterized with the change in (scaled) log-likelihood $L(\px)$ given the observed empirical distribution $\tau$.
It sufficed then to show that Jeffrey's update rule increases the likelihood, concluding the main proof.

The proof for the latter decomposes the log-likelihood into two separate functions, one ($H$) which is non-negative due to Gibb's inequality, and another one ($Q$) which is a function that is maximal at Jeffrey's posterior.

\bibliographystyle{plain}
\bibliography{bibliography}

\begin{thebibliography}{1}

\bibitem{dempster1977maximum}
Arthur~P Dempster, Nan~M Laird, and Donald~B Rubin.
\newblock Maximum likelihood from incomplete data via the em algorithm.
\newblock {\em Journal of the royal statistical society: series B (methodological)}, 39(1):1--22, 1977.

\bibitem{elsalamouny2020generalized}
Ehab ElSalamouny and Catuscia Palamidessi.
\newblock Generalized iterative bayesian update and applications to mechanisms for privacy protection.
\newblock In {\em 2020 IEEE European Symposium on Security and Privacy (EuroS\&P)}, pages 490--507. IEEE, 2020.

\bibitem{jacobs2021learning}
Bart Jacobs.
\newblock Learning from what's right and learning from what's wrong.
\newblock {\em arXiv preprint arXiv:2112.14045}, 2021.

\bibitem{mclachlan2007algorithm}
Geoffrey~J McLachlan and Thriyambakam Krishnan.
\newblock {\em The EM algorithm and extensions}.
\newblock John Wiley \& Sons, 2007.

\bibitem{wu1983convergence}
CF~Jeff Wu.
\newblock On the convergence properties of the em algorithm.
\newblock {\em The Annals of statistics}, pages 95--103, 1983.

\end{thebibliography}

\end{document}